\documentclass[10pt,twocolumn,letterpaper]{article}

\usepackage{cvpr}
\usepackage{times}
\usepackage{epsfig}
\usepackage{graphicx}
\usepackage{amsmath}
\usepackage{amssymb}
\usepackage{multirow}
\usepackage{array}
\usepackage{caption}
\usepackage{subcaption}
\usepackage{pifont}
\usepackage{url}
\newcommand{\cmark}{\ding{51}}%
\newcommand{\xmark}{\ding{55}}%

\DeclareMathOperator*{\argmin}{arg\,min}

\usepackage[breaklinks=true,colorlinks,bookmarks=false]{hyperref}

\cvprfinalcopy 

\def\cvprPaperID{7685} 

\ifcvprfinal\fi
\begin{document}

\title{SESS: Self-Ensembling Semi-Supervised 3D Object Detection}

\author{Na Zhao \quad Tat-Seng Chua \quad Gim Hee Lee \\
	Deaprtment of Computer Science, National University of Singapore\\
	{\tt\small \{nazhao, chuats, gimhee.lee\}@comp.nus.edu.sg}
}

\maketitle
\thispagestyle{empty}

\begin{abstract}
	The performance of existing point cloud-based 3D object detection methods heavily relies on large-scale high-quality 3D annotations. However, such annotations are often tedious and expensive to collect. Semi-supervised learning is a good alternative to mitigate the data annotation issue, but has remained largely unexplored in 3D object detection. Inspired by the recent success of self-ensembling technique in semi-supervised image classification task, we propose SESS, a self-ensembling semi-supervised 3D object detection framework. Specifically, we design a thorough perturbation scheme to enhance generalization of the network on unlabeled and new unseen data.
	Furthermore, we propose three consistency losses to enforce the consistency between two sets of predicted 3D object proposals, to facilitate the learning of structure and semantic invariances of objects. Extensive experiments conducted on SUN RGB-D and ScanNet datasets demonstrate the effectiveness of SESS in both inductive and transductive semi-supervised 3D object detection. Our SESS achieves competitive performance compared to the state-of-the-art fully-supervised method by using only 50\% labeled data. Our code is available at \url{https://github.com/Na-Z/sess}.
\end{abstract}

\section{Introduction}
\begin{figure}[t]
	\centering
	\includegraphics[scale=0.28]{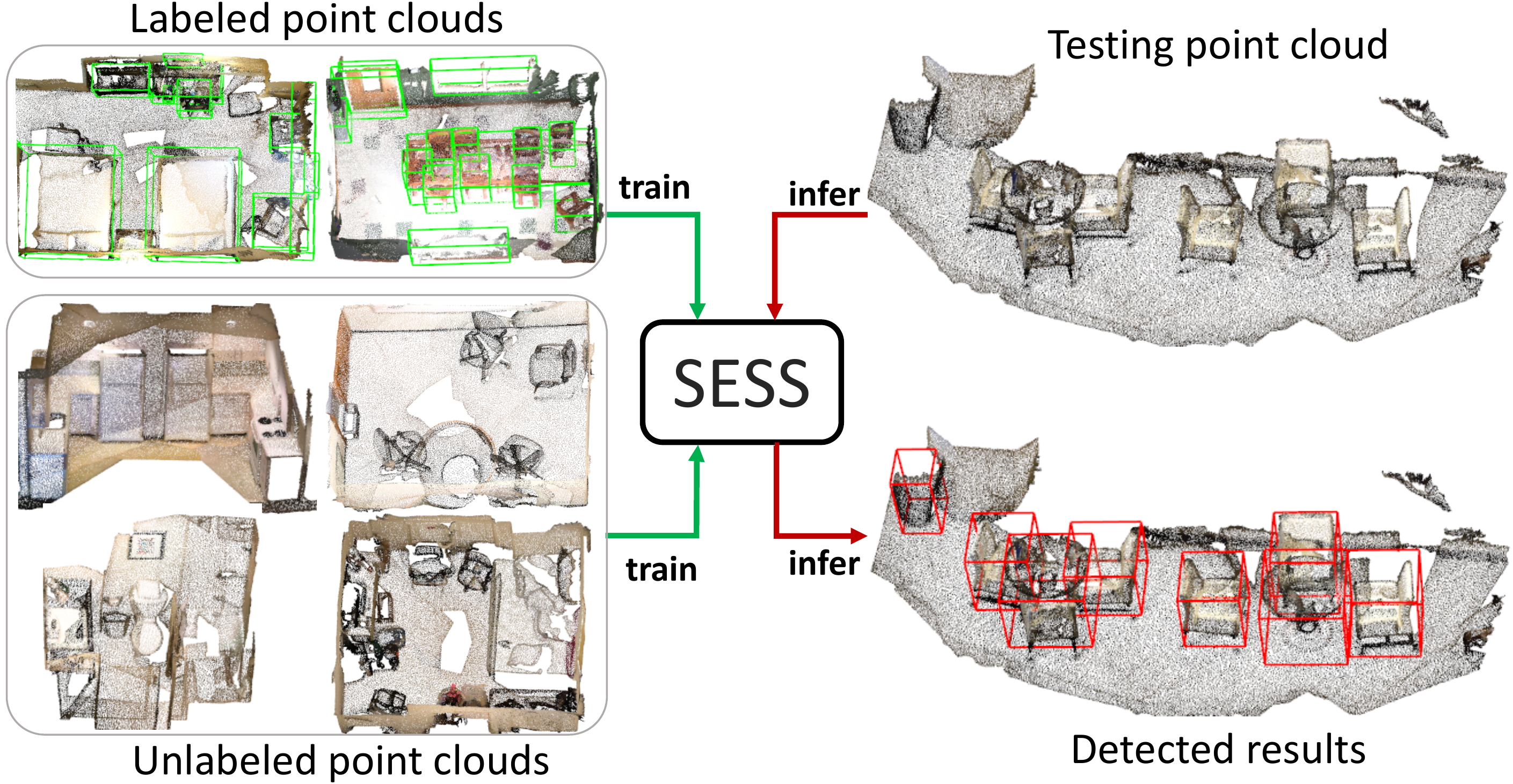}
	\caption{\small{\textbf{Semi-supervised 3D object detection pipeline}. Our SESS can predict 3D bounding boxes and semantic labels of objects for an unlabeled scene
			after training with a mixture of labeled data and unlabeled data.}}
	\label{fig:teaser}
	\vspace{-0.1in}
\end{figure}

Point cloud-based 3D object detection is the task to estimate the object category and oriented 3D bounding box for all objects in the scene. This task has always been a great interest to computer vision and robotics communities due to its potential real-world applications in many areas such as autonomous driving, domestic robotics, augmented/virtual reality, etc. In recent years, many deep learning-based approaches for point cloud-based 3D object detection~\cite{chen2019fast, lang2019pointpillars, liang2019multi, qi2019deep, qi2018frustum, ren20183d, shi2019pointrcnn, simon2018complex, wang2019frustum, yang2019std, zhou2018voxelnet} have emerged and achieved high performances on various benchmark datasets~\cite{dai2017scannet, Geiger2012CVPR, song2015sun}. Despite the impressive performances, most of the existing deep learning-based approaches for 3D object detection on point clouds are strongly supervised and require the availability of a large amount of well-annotated 3D data that is often time-consuming and expensive to collect. 

Semi-supervised learning is a promising alternative to strongly supervised learning for point cloud-based 3D object detection. This is because semi-supervised learning requires only few labeled data, and this largely alleviates the difficulty to collect enormous amount of labeled data. Furthermore, the available few strong labels can still provide the necessary supervision to guide the deep network into learning the correct information for 3D object detection. Information from the few strong labels can also be propagated to the unlabeled data to improve learning.
A complete removal of strong labels in the training data would be extremely challenging for the deep network to learn anything meaningful. This is due to the inherent difficulty for a deep network to precisely detect 3D bounding boxes of objects in the point cloud, where points are sparsely distributed, and/or the scene is partially visible and incomplete due to occlusions and 3D amodal perception.
To the best of our knowledge, \cite{tang2019transferable} is currently the only existing work to learn a deep network for point cloud-based 3D object detection without strong supervision. More specifically, they propose a cross-category semi-supervised learning where 3D ground truth labels are needed for a set of object classes, and 2D ground truth labels are required for all object classes. Although promising results are achieved in \cite{tang2019transferable}, the approach requires RGB-D input and does not work on pure 3D point clouds. Moreover, it still requires a large amount of 3D labels on the strong object classes. 

In view of the potential of semi-supervised learning and limitations in \cite{tang2019transferable}, we address the in-category semi-supervised 3D object detection problem with 3D point cloud as the only input in this paper. In contrast to cross-category semi-supervision, in-category semi-supervision means that the training data contains few strongly labeled point clouds and a large number of unlabeled point clouds. Furthermore, the strongly labeled point clouds are assumed to contain all object classes of interests, albeit few examples per object class. To this end, we propose SESS - a \textbf{S}elf-\textbf{E}nsembling \textbf{S}emi-\textbf{S}upervised 3D object detection framework for point clouds. More specifically, our SESS achieves semi-supervision with a Mean Teacher paradigm~\cite{tarvainen2017mean} that contains a teacher and student 3D object detection network. The teacher guides the predictions of the student to be consistent with its predictions under random perturbations, where these predictions are sets of 3D object proposals. In other words, we want the 3D object proposals from both teacher and student networks to be aligned at the end of the training stage. We propose three consistency losses  based on the center, class and size of the 3D object proposals to encourage alignment of the 3D object proposals from the teacher and student networks. Our three consistency losses encode both geometry and semantic information to guide the network towards learning precise coordinates of the 3D bounding boxes and accurate object categories. We conduct experiments of our SESS framework on two benchmark datasets. Promising results over baseline and strongly supervised approaches validate our semi-supervised learning approach for the challenging task of point cloud-based 3D object detection in both inductive and transductive semi-supervised learning settings. 


\section{Related work}
\subsection{3D Object Detection}
A number of approaches have been proposed for 3D object detection task, which can be briefly summarized into three different types based on their input data formats: 2D projection~\cite{li2019gs3d, liang2019multi, simon2018complex, yang2018pixor}, voxel grid~\cite{chen2019fast, lang2019pointpillars, ren2016three, song2016deep, ren20183d, yan2018second, zhou2018voxelnet}, and point cloud~\cite{lahoud20172d, qi2019deep, qi2018frustum, shi2019pointrcnn, wang2019frustum, yang2019std, yi2019gspn}. 
The 2D projection and voxel grid based methods are proposed to circumvent the difficulty in processing irregular point clouds by either projecting 3D data into 2D representations (\eg front-view, or bird's eye view) or voxelizing it into regular grids. To efficiently localize 3D objects in the point cloud of a 3D space, \cite{lahoud20172d, qi2018frustum,wang2019frustum} leverage on mature 2D object detectors to trim a 3D bounding frustum for each detected object, for 3D search space reduction, while \cite{qi2019deep, shi2019pointrcnn, yang2019std, yi2019gspn} explore the sparsity of 3D data and generate 3D proposals around seed points that are determined by different manners (\eg segmenting~\cite{shi2019pointrcnn} or voting~\cite{qi2019deep}). 
Despite the significant improvement achieved by the existing detection models, a large number of high-quality 3D ground truths are required for training. This limits their applicability in practice, where the ground truths are expensive to acquire. 

In order to alleviate the limitation and leverage the abundant unlabeled data that are easier to access, semi-supervised 3D object detection is a promising direction to exploit. However, there is no existing semi-supervised point cloud-based 3D object detection approach that only involves a small set of labeled data. The most closely related work is proposed recently by Tang and Lee in \cite{tang2019transferable}. They propose a cross-category semi-supervised 3D object detection method. However, it requires all the 2D box labels and some of the 3D box labels. We consider this setting as ``mix supervised'' to differentiate with our semi-supervised setting where few labeled samples are used with plentiful of unlabeled samples. Furthermore, \cite{tang2019transferable} follows the two-step pipeline in~\cite{qi2018frustum} to restrict the object localizing space: the first step is 2D object detection on RGB images and the second step is 3D object detection in the frustum point clouds yielded from the 2D detections. This two-step pipeline means that the performance is tightly dependent on the performance of the 2D detector. In this work, we directly process the raw point cloud in one step to remove the dependency on 2D modality.

\begin{figure*}[t]
	\centering
	\includegraphics[scale=0.55]{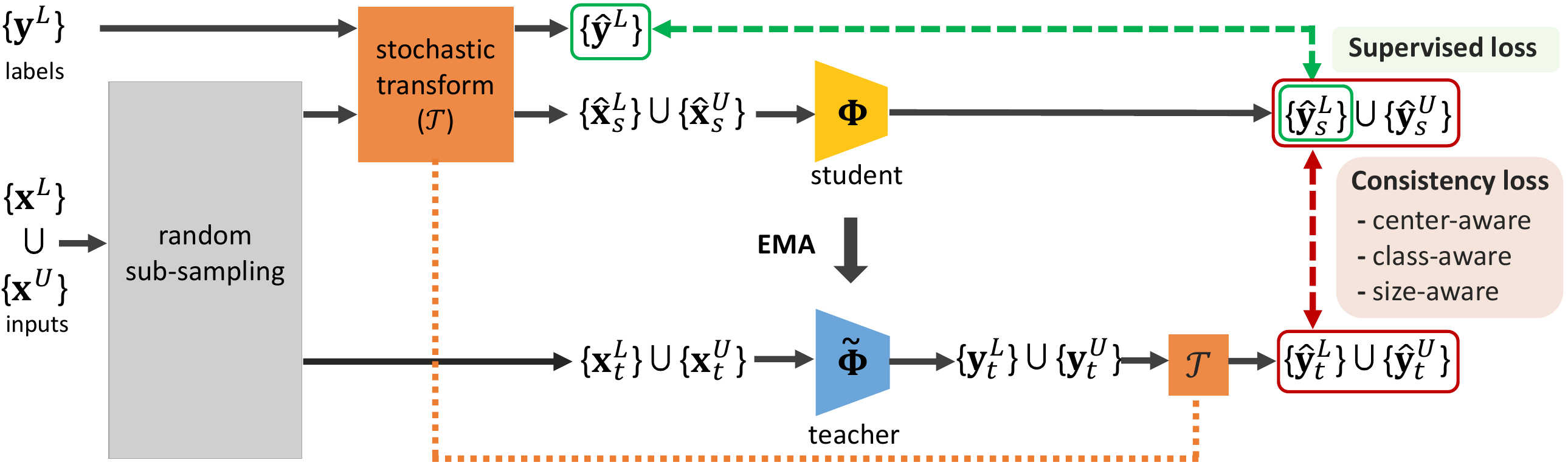}
	\caption{\small{The architecture of our SESS. In this figure, a training batch, including a set of labeled sample $\{\textbf{x}^L\}$ and a set of unlabeled samples $\{\textbf{x}^U\}$, is passed through different perturbations and then input into the student and the teacher network, respectively. The predictions of the student network are compared with the corresponding ground truth labels $\{\textbf{y}^L\}$ processed by the same transformation $\mathcal{T}$ using supervised loss and with the teacher predictions processed by the same transformation $\mathcal{T}$ using consistency loss.}}
	\label{fig:OurNetwork}
\end{figure*}

\subsection{Semi-Supervised Learning}
Semi-Supervised Learning (SSL) attracts growing interest in a wide range of research areas (\eg image classification and segmentation) by virtue of its aim to learn from both labeled and unlabeled data simultaneously. Many approaches have been proposed to solve SSL. Due to the space limitation, we only review self-ensembling based approaches, the most promising direction in SSL recently. 

The idea behind self-ensembling approaches is to improve the generalization of a model by encouraging consensus among ensemble predictions of unknown samples under small perturbations of inputs or network parameters.   
For instance, $\Gamma$ model~\cite{rasmus2015semi}, a variation of ladder network~\cite{valpola2015neural}, consists of two identical parallel branches that respectively take one image and the corrupted version of the image as input. The consistency loss is computed based on the difference between the (pre-activated) predictions from the clean branch and the (pre-activated) corrupted branches processed by an explicit denoising layer.
In contrast to $\Gamma$ model, $\Pi$ model~\cite{laine2017temporal} discards the explicit denoising layer and inputs the same image with different corruption conditions into a single branch. Virtual Adversarial Training~\cite{miyato2018virtual} shares similar idea with the $\Pi$ model but uses adversarial perturbation instead of independent noise.
Temporal model~\cite{laine2017temporal}, an extension of $\Pi$ model, forces the consistency between the recent network output and the aggregation of network predictions over multiple previous training epochs rather than predictions from auxiliary corrupted input. However, this model becomes cumbersome when applied to large dataset because it needs to maintain a per-sample moving average of the historical network predictions.  Mean Teacher~\cite{tarvainen2017mean} tackles the weakness of temporal model by replacing network prediction average with network parameter average. It contains two network branches - \textit{teacher} and \textit{student} with the same architecture. The parameters of the teacher are the exponential moving average of the student network parameters that are updated by stochastic gradient descent. The student network is trained to yield consistent predictions with the teacher network. We choose the Mean Teacher paradigm as the basis of our framework, and adapt it to the 3D object detection task.

\section{Our Method}
\subsection{Problem Definition}
Given any point cloud of a scene as input, our objective is to classify and localize amodal 3D bounding boxes for objects in the 3D scene.
In the semi-supervised setting, we have access to $N$ training samples, including $N_l$ labeled point clouds $\mathcal{P}^L = \{\textbf{x}_i^L, \textbf{y}_i^L\}_{i=1}^{N_l}$ and $N_u$ unlabeled point clouds $\mathcal{P}^U = \{\textbf{x}_i^U\}_{i=1}^{N_u}$. Here $\textbf{x}_i \in \mathbb{R}^{n \times 3}$ denotes the point cloud of a 3D scene, containing $n$ points with coordinates; and $\textbf{y}_i^L$ denotes the ground truth annotations for all the interested objects in the 3D point cloud $\textbf{x}_i^L$. Each object is represented by a semantic class $s$ (1-of-$K$ predefined classes) and an amodal 3D bounding box parameterized by its center $c = (c^x, c^y, c^z )$, size $d = (l, w, h)$, and orientation $\theta$ along the upright-axis. 

\subsection{SESS Architecture}
The illustration of our SESS architecture is shown in Figure~\ref{fig:OurNetwork}. We use the Mean Teacher paradigm~\cite{tarvainen2017mean} in our semi-supervised 3D object detection task, where the student and the teacher networks are 3D object detectors. 
The student and teacher networks take the perturbed point clouds as input and output the 3D object proposals, which represent the estimated classes and 3D bounding boxes of all the objects of interest in the point cloud. We adopt the state-of-the-art VoteNet\footnote{It is worth highlighting that instead of designing a specific detector model, our proposed framework is model-agnostic and any existing point cloud-based 3D object detection network can be used.}~\cite{qi2019deep} as our backbone for the student and teacher networks. 
More specifically, SESS takes a training batch with a mixture of labeled and unlabeled point clouds: $\{\textbf{x}_i^L\}_{i=1}^{B_l} \cup \{\textbf{x}_i^U\}_{i=1}^{B_u}$, where $B_l$ and $B_u$ denote the labeled and unlabeled samples in a batch, respectively. We randomly sample $M$ points from each training point cloud, \ie $\textbf{x}^L$ or $\textbf{x}^U$, twice to get two sets of points. The first set of points $\textbf{x}_s$ is perturbed into $\hat{\textbf{x}}_s$ by a stochastic transformation $\mathcal{T}$ and then passed to the student network, while the second set of points $\textbf{x}_t$ is directly passed to the teacher network. The output proposals from the teacher network $\textbf{y}_t$ are further transformed to $\hat{\textbf{y}}_t$ by the $\mathcal{T}$ applied on $\textbf{x}_s$ previously.
For each proposal in $\hat{\textbf{y}}_t$, we find its closest alignment from the output proposals of the student network $\hat{\textbf{y}}_s$ based on the Euclidean distance. Subsequently, the error between each aligned proposal pair is computed from three consistency losses. Concurrently, the set of ground truths $\textbf{y}^L$ is also transformed by the same $\mathcal{T}$ applied on $\textbf{x}^L_s$, and the transformed $\hat{\textbf{y}}^L$ is compared with the labeled output of the student network $\hat{\textbf{y}}_s^L$ using a supervised loss.
Finally, the parameters of the student network $\Phi$ is updated via gradient descent at training step $t$, and then 
the updated parameters from the student network are used in an exponential moving average (EMA) to update the
parameters of the teacher network $\tilde{\Phi}$:
\vspace{-0.15in}
\begin{equation}
	\tilde{\Phi}_{t} = \alpha \tilde{\Phi}_{t-1}  + (1-\alpha)\Phi_{t},
\end{equation}
where $\alpha$ is a smoothing hyper-parameter that controls how much information the teacher takes from the student network.
For supervised loss, we take the same multi-task loss as in \cite{qi2019deep}. We will introduce our perturbation scheme and consistency losses for adapting the Mean Teacher paradigm into the 3D object detection task below.

\subsection{Perturbation Scheme}
As mentioned in \cite{laine2017temporal, tarvainen2017mean}, input perturbation or data augmentation play an essential role in the success of self-ensembling approaches. The perturbation schemes of the Mean Teacher on image-based tasks, \eg image recognition, include random translations and horizontal flips of the input images, adding Gaussian noises on the input layer, and applying dropouts within the network. However, none of the image-based perturbation schemes can be used directly for our point cloud-based 3D object detection task. Consequently, 
we propose a perturbation scheme suitable for point cloud-based 3D object detection in this paper.
\vspace{-0.1in}
\paragraph{Random Sub-sampling.} We apply random sub-sampling on the input point cloud to both the student and teacher networks as part of our perturbation scheme. The local geometrical relationship of the points in two random sub-samples of a given point cloud might differ significantly, but the global geometry, \ie the 3D bounding box locations of the objects, in the sub-sampled point clouds should remain the same. As a result, 
our model is trained to exploit the underlying geometry in the global context by forcing the consistency between the stochastic outputs from the student and teacher networks. 
\vspace{-0.1in}
\paragraph{Stochastic Transform.}
We apply stochastic transformations that include flipping, rotation and scaling on the randomly sub-sampled point cloud for the student network to prevent the network from memorizing unintended properties of the training point clouds, \eg the absolute position of each point. More specifically, 
we formulate the transformation operations as a set of stochastic variables $\mathcal{T}=\{\mathcal{F}_x, \mathcal{F}_y, \mathcal{R},  \mathcal{S} \}$. Here $\mathcal{F}_x$ represents a random flip along the $x$-axis, and its binary value is determined by: 
\vspace{-0.1in}
\begin{equation}
	\mathcal{F}_x =
	\begin{cases}
		1 & if ~ \epsilon > 0.5, \\
		0 & otherwise, \\
	\end{cases} 
\end{equation} 
where $\epsilon$ is a random variable uniformly sampled from $[0,1]$. $\mathcal{F}_y$ represents a random flip along the $y$-axis and is generated the same way as $\mathcal{F}_x$. $\mathcal{R}$ denotes the rotation around the upright-axis, paramterized by a rotation angle $\omega$ sampled uniformly from $[-\vartheta,+\vartheta]$:
\vspace{-0.1in}
\begin{equation}
	\mathcal{R(\omega)} = 
	\begin{bmatrix}
		\cos(\omega) &-\sin(\omega) & 0 \\
		\sin(\omega)  & \cos(\omega) & 0 \\
		0 & 0 & 1
	\end{bmatrix}.
\end{equation}
And $\mathcal{S}$ that is uniformly sampled from $[a,b]$ represents the scaling of the points.
Finally, a $\mathcal{T}_i$ is randomly sampled and applied on each input training point cloud $\textbf{x}_s$ to the student network as:
$\hat{\textbf{x}}_s = \mathcal{T}_i * \textbf{x}_s$. Note that the ground truth labels $\textbf{y}_i^L$ of the labeled input point cloud $\textbf{x}_i^L$ are also transformed by the corresponding $\mathcal{T}_i$ before computing the supervised loss.
Additionally, the output proposals $\textbf{y}_t$ from the teacher network are also transformed by $\mathcal{T}_i$ to enable the alignment between outputs of the two networks. 

\subsection{Consistency Loss}
Unlike the direct computation of consistency between class predictions of perturbed images in the context of recognition task~\cite{tarvainen2017mean}, the consistency between two sets of 3D object proposals cannot be computed directly. 
We circumvent this problem by pairing up the predicted proposals from the student and teacher networks with an alignment scheme, followed by applying three consistency losses on the paired proposals. 
The objective of the three consistency losses is to enforce the consensus of object locations, semantic categories and sizes.
Let $\hat{C}_s=\{\hat{c}_s\}$ denotes the centers of the predicted 3D bounding boxes from the student network, and $\hat{C}_t = \{\hat{c}_t\}$ denotes those from the teacher network after transformation. 
For each $\hat{c}_t \in \hat{C}_t$, we do the alignment by searching for the its nearest neighbor in $\hat{C}_s$ based on the minimum Euclidean distance between the centers of the bounding boxes. We further use $\hat{C}_s^{\mathcal{A}}$ to denote the elements from $\hat{C}_s$ that are aligned with each element in $\hat{C}_t$. More formally,
\vspace{-0.05in}
\begin{align}
	&\hat{C}^{\mathcal{A}}_s = \{ \cdots, \hat{c}^{\mathcal{A}}_{s_j}, \cdots \} : \notag \\
	&\hat{c}^{\mathcal{A}}_{s_j} = \argmin_{\hat{c}_s} \left\Vert \hat{c}_s - \hat{c}_{t_j}  \right\Vert_2,~\forall \hat{c}_s  \in \hat{C}_s.
\end{align}
Similarly, we can also collect $\hat{C}^{\mathcal{A}}_t$ with elements from $\hat{C}_t$ that are aligned with each element in $\hat{C}_s$. It is important to note that the alignments $\hat{C}^{\mathcal{A}}_s$ and $\hat{C}^{\mathcal{A}}_t$ are not bijective, hence $\hat{C}^{\mathcal{A}}_s \neq \hat{C}^{\mathcal{A}}_t$.
Intuitively, the alignment errors, \ie, the total distance between all corresponding elements in $\hat{C}^{\mathcal{A}}_s \leftrightarrow \hat{C}_t$ and $\hat{C}^{\mathcal{A}}_t \leftrightarrow \hat{C}_s$, should be zero when the bounding boxes predicted by the teacher and student networks are consistent. Thus, we propose the \textbf{center-aware consistency loss}:
\vspace{-0.05in}
\begin{equation}
	\mathcal{L}_{center} =\frac{ \sum_{\hat{c}_s} \| \hat{c}_s - \hat{c}_t^{\mathcal{A}} \|_2  + \sum_{\hat{c}_t} \| \hat{c}_t - \hat{c}_s^{\mathcal{A}} \|_2 }{|\hat{C}_s|+|\hat{C}_t|},
\end{equation} 
to minimize the alignment errors between the teacher and student network.

In addition to center consistency, we also consider two other properties of the 3D proposals: semantic class and size to enforce the consistency between two sets of proposals.
Following the principle in classic self-ensembling learning, where the teacher network produces targets for the student to learn, we only consider a uni-directional alignment, \ie, $\hat{C}_t$ to $\hat{C}_s^{\mathcal{A}}$ in computing the class- and size-aware consistency losses. 
More specifically, let $\hat{P}_s = \{\hat{p}_s\}$ and $\hat{P}_t = \{\hat{p}_t\}$ denote the class probabilities of the predicted objects from the student and the teacher network, respectively. The aligned $\hat{P}_s^{\mathcal{A}} = \{ \hat{p}_s^{\mathcal{A}}\}$ is easily obtained based on minimum center distance.  
We define the \textbf{class-aware consistency loss} as the Kullback-Leibler (KL) divergence between $\hat{P}_s^{\mathcal{A}}$ and $\hat{P}_t$:
\vspace{-0.1in}
\begin{equation}
	\mathcal{L}_{class} = \frac{1}{|\hat{P}_t|} \sum D_{KL} (\hat{p}_s^{\mathcal{A}} \parallel \hat{p}_t).
\end{equation} 
In similar vein, the sizes of the bounding boxes predicted by the student and the teacher networks are denoted as $\hat{D}_s = \{\hat{d}_s\}$ and $\hat{D}_t = \{\hat{d}_t\}$, respectively. 
We use the same minimum center distance to get the aligned $\hat{D}_s^{\mathcal{A}} = \{ \hat{d}_s^{\mathcal{A}}\}$.
The \textbf{size-aware consistency loss} can now be computed as the Mean Square Error (MSE) between $\hat{D}_s^{\mathcal{A}}$ and $\hat{D}_t$:
\vspace{-0.05in}
\begin{equation}
	\mathcal{L}_{size} = \frac{1}{|\hat{D}_t|} \sum (\hat{d}_s^{\mathcal{A}} - \hat{d}_t)^2.
\end{equation} 
Finally, the total consistency loss is a weighted sum of all the three consistency terms described earlier:
\vspace{-0.05in}
\begin{equation}
	\mathcal{L}_{consistency} = \lambda_1 \mathcal{L}_{center} + \lambda_2 \mathcal{L}_{class} + \lambda_3 \mathcal{L}_{size},
\end{equation} 
where $\lambda_1$,  $\lambda_2$, and  $\lambda_3$ are the weights to control the importance of the corresponding consistency term.


\begin{table*}[h]
	\centering
	\caption{\small{Comparison with VoteNet on SUN RGB-D val set and ScanNetV2 val set with varying ratios of labeled data. mAP@0.25 are reported as mean$\pm$standard deviation, based on 3 runs with random sampling. And the improvement (Improv.) is computed based on the mean performances over 3 runs. Note that our SESS is initialized by the VoteNet weights pre-trained on the corresponding labeled data.}}
	\scalebox{0.95}{
		\begin{tabular}{l ||c|c c c c c c  c}
			\hline
			Dataset & Model  & 10\% & 20\% & 30\% & 40\% & 50\% & 70\% & 100\% \\
			\hline
			\multirow{2}{*}{\small{SUNRGB-D}}
			& VoteNet~\cite{qi2019deep} & 34.43$\pm$1.07  & 41.13$\pm$0.36 & 47.70$\pm$0.17 & 50.77$\pm$0.19 & 52.5$\pm$0.19 & 56.13$\pm$0.18 & 57.7 \\ \cline{2-9}
			&\textbf{SESS} & \textbf{42.87$\pm$1.01} & \textbf{47.87$\pm$0.48} &\textbf{53.17$\pm$0.63} & \textbf{54.73$\pm$0.26} & \textbf{56.37$\pm$0.22}  & \textbf{58.97$\pm$0.17} & \textbf{61.1} \\
			&Improv.(\%) & 24.51$\uparrow$ & 16.39$\uparrow$& 11.47$\uparrow$& 7.80$\uparrow$& 7.37$\uparrow$ & 5.06$\uparrow$ & 5.89$\uparrow$\\
			\hline \hline
			\multirow{2}{*}{ScanNetV2}
			& VoteNet~\cite{qi2019deep} & 30.97$\pm$0.79 & 41.60$\pm$0.46 & 45.57$\pm$0.38 & 49.2$\pm$0.33 & 52.57$\pm$0.07 & 54.97$\pm$0.07 & 58.6 \\ \cline{2-9}
			&\textbf{SESS} & \textbf{39.67$\pm$0.91} & \textbf{47.93$\pm$0.39} &  \textbf{52.20$\pm$0.09} & \textbf{54.93$\pm$0.27} & \textbf{57.77$\pm$0.41} & \textbf{59.20$\pm$0.08} & \textbf{62.1}\\
			&Improv.(\%) & 28.09$\uparrow$ & 15.22$\uparrow$  & 14.55 $\uparrow$ & 11.64 $\uparrow$ & 9.89 $\uparrow$ & 7.70 $\uparrow$ & 5.97$\uparrow$\\
			\hline
		\end{tabular}
	}
	\label{tbl:semisupervise-comparison}
\end{table*}

\begin{table}[h]
	\centering
	\caption{\small{Comparison with fully-supervised methods on SUN RGB-D and ScanNetV2 val sets with 100\% training labels.}}
	\scalebox{1}{
		\begin{tabular}{l ||c c } \hline
			Method & SUN RGB-D & ScanNetV2 \\ \hline
			DSS~\cite{song2016deep} & 42.1 & 15.2 \\ 
			COG~\cite{ren2016three} & 47.6 & -- \\ 
			2D-driven~\cite{lahoud20172d} & 45.1 & -- \\
			F-PointNet~\cite{qi2018frustum} & 54.0 &  19.8  \\ 
			GSPN~\cite{lahoud20172d} & -- &30.6  \\ 
			3D-SIS~\cite{hou20193d} & -- & 40.2 \\
			VoteNet~\cite{qi2019deep} & 57.7 & 58.6 \\ \hline
			\textbf{SESS} & \textbf{61.1} & \textbf{62.1} \\ 
			\hline
		\end{tabular}
	}\label{tbl:fullysupervise-comparison}
\end{table}

\section{Experiments}
\subsection{Datasets}
We evaluate our SESS on SUN RGB-D and ScanNet for semi-supervised 3D object detection.
\vspace{-0.1in}
\paragraph{SUN RGB-D}\cite{song2015sun} is an indoor benchmark dataset for 3D object detection. It contains 10,335 single-view RGB-D images, which are officially split into 5,285 training samples and 5,050 validation samples, where 3D bounding box annotations for hundreds of object classes are available. Followed the standard evaluation protocol~\cite{lahoud20172d, qi2019deep, qi2018frustum, ren2016three, tang2019transferable}, we perform evaluation on the 10 most common categories for comparing with the previous methods. By using the provided camera parameters, the depth images are converted to point clouds as our inputs.
\vspace{-0.1in}
\paragraph{ScanNetV2}\cite{dai2017scannet} contains 1,513 reconstructed meshes from 707 unique indoor scenes, which are officially split into 1,201 training samples and 312 validation samples. Each scene is well annotated with semantic segmentation masks. Since there is no existing amodal or orientated 3D bounding box in ScanNetV2 dataset, we derive the axis-aligned bounding boxes from the point-level labeling as in~\cite{hou20193d, qi2019deep}. We adopt the same 18 object classes out of the 21 semantic classes as proposed in~\cite{hou20193d, qi2019deep}. The input point clouds are generated by sampling vertices from meshes.

For both datasets, we evaluate on different proportions of labeled data randomly sampled from all the training data. We ensure that all classes are present, or otherwise we re-sample until all the $K$ classes are covered in the labeled set.  
We keep the remaining data as unlabeled data for training in our semi-supervised framework.  

\subsection{Implementation Details}
\paragraph{Framework Details.}
We feed training batches of point clouds with 50,000 points to our framework. To construct a batch, we randomly sample $B_l$ labeled samples from $\mathcal{P}^L$ and $B_u$ unlabeled samples from $\mathcal{P}^U$. In the experiments, $B_l$ is set to 2 and $B_u$ to 8. During the perturbation step, the number of randomly sub-sampled points is 40,000; the $\vartheta$ is set to 30$^{\circ}$ on SUN RGB-D and 5$^{\circ}$ on ScanNetV2; the random scale range is bounded by $a=0.85$ and $b=1.15$. The weights in the consistency loss function are set as $\lambda_1=1$, $\lambda_1=2$, $\lambda_3=1$. As suggested in \cite{tarvainen2017mean}, we ramp up the coefficient of consistency cost from 0 to its maximum value of 10 during the first 30 epochs, using a sigmoid-shaped function $e^{-5(1-T)^2}$, where $T$ increases linearly from 0 to 1 during the ramp-up period.
In terms of EMA decay $\alpha$, we set $\alpha=0.99$ during the ramp-up period, and $\alpha=0.999$ for the rest of the training, following \cite{tarvainen2017mean}.
\vspace{-0.1in}
\paragraph{Training.}
We adopt the exact network structure of VoteNet \cite{qi2019deep} as the structure of our student and teacher network. We pre-train VoteNet with all the available labeled samples. We then initialize the student and teacher networks with the pre-trained weights, and train the student network on both the labeled and unlabeled data by minimizing the supervised loss as well as consistency loss. The student network is trained by an ADAM optimizer with an initial learning rate of 0.001. The learning rate is decayed by 0.1 at the 80$^{th}$ epoch. In general, the model converges at around 100 epochs. The number of generated 3D proposals is 128.
\vspace{-0.1in}
\paragraph{Inference.}
During inference, we forward the point cloud of a scene to the student network\footnote{Note that the teacher network can also be used to detect objects. In the experiments, we find the student and the teacher network give similar performance.} to generate the proposals. Following the same protocol as described in \cite{qi2019deep}, we post-process those predicted proposals by a 3D NMS module with an 3D Intersection-over-Union(IoU) threshold of 0.25. For the evaluation metric, we adopt the widely used mean average precision (mAP). By default, mAP@0.25 (3D IoU threshold 0.25) is reported in the following experiments.

\subsection{Comparison with Fully-supervised Methods}
\paragraph{Baselines.}
To the best of our knowledge, there are no other 3D object detection approaches sharing the same semi-supervised setting as us. Hence, we compare our semi-supervised SESS to the state-of-the-art fully-supervised 3D object detection method, VoteNet~\cite{qi2019deep}, which can be considered as an upper bound of our semi-supervised method since we share the same network backbone. By drawing varying ratios of labeled data out of the entire training set, we train VoteNet with the available labeled data in a fully-supervised way, and SESS with the available labeled data as well as the remaining unlabeled data in a semi-supervised way.
Additionally, we also evaluate our semi-supervised SESS based on a wide-ranging comparison with existing fully-supervised 3D object detection methods. Deep Sliding Shapes (DSS)~\cite{song2016deep} and Cloud of gradients (COG)~\cite{ren2016three} are both sliding window based methods, where DSS is a 3D extension of Faster R-CNN pipeline~\cite{ren2015faster}, and COG designs a 3D HoG-like feature to model the 3D geometry and appearance. 2D-driven~\cite{lahoud20172d} and F-PointNet~\cite{qi2018frustum} both depend on 2D detection in associated RGB images to reduce the search space of 3D localization. GSPN~\cite{lahoud20172d} and 3D-SIS~\cite{hou20193d} both target on 3D instance segmentation task but incorporate 3D object detection as an auxiliary task. 
Note that all the aforementioned methods use both point clouds and RGB images as inputs except VoteNet and our SESS that only require point clouds.

\vspace{-0.1in}
\paragraph{Results.} Table \ref{tbl:semisupervise-comparison} lists the comparison results against VoteNet under different ratios of labeled data on the two datasets, respectively. SESS significantly outperforms VoteNet under each ratio setting. The improvements verify the effectiveness of our proposed semi-supervised framework. On both datasets, as the proportion of labeled samples decreases, the performance gap between our SESS and the fully-supervised VoteNet becomes larger. Given 10\% labeled data, our SESS gains around 24.51\% and 28.09\% improvement over VoteNet on SUN RGB-D and ScanNetV2, respectively. This indicates that our framework is able to learn knowledge from unlabeled data, and our benefit is larger when the number of labeled data is scarce. 

It is interesting to see that by using only 50\% labeled samples, our SESS achieves close to the upper-bound performance obtained by the fully-supervised VoteNet with 100\% labeled samples on both datasets.
Furthermore, it is worth pointing out that when given all the labeled training data, our SESS is able to further improve the performance beyond the upper-bound performance of VoteNet. 
We attribute the outperformance of SESS to its consistency regularization mechanism, where the 3D detector is trained to be robust to various perturbations, and the proposed three consistency losses that encode both geometry and semantic information guide the 3D detector towards producing more accurate predictions.
This further indicates that our consistency losses are complementary to supervised loss, and our framework might be integrated with any supervised 3D object detector to enhance the detection accuracy.

In Table \ref{tbl:fullysupervise-comparison}, we further list the performance comparison between SESS and various fully-supervised methods on the two datasets, by using all the training samples. 

\begin{table}[t]
	\centering
	\caption{\small{Transductive leaning on SUN RGB-D and ScanNetV2 unlabeled training sets, compared with fully-supervised VoteNet. The percentage indicates the ratio of labeled data for training.}}
	\scalebox{0.83}{
		\begin{tabular}{l ||l|l l l l l l}
			\hline
			Dataset & Model  & 10\% & 20\% & 30\% & 40\% & 50\% & 70\%  \\
			\hline
			\multirow{2}{*}{\small{SUNRGB-D}}
			& VoteNet & 33.5 & 39.8 & 47.5 & 49.7 & 51.6 & 55.2\\ 
			&\textbf{SESS} & \textbf{40.7} & \textbf{46.1} & \textbf{53.3} & \textbf{54.3} & \textbf{55.1} & \textbf{59.0}\\
			\hline \hline
			\multirow{2}{*}{ScanNetV2}
			& VoteNet & 37.8 & 47.7 & 52.1 & 56.9 & 61.2 & 64.3\\ 
			&\textbf{SESS} & \textbf{46.7} & \textbf{55.4} &  \textbf{59.5} & \textbf{63.9} & \textbf{67.5} & \textbf{69.6}\\
			\hline
		\end{tabular}
	}\label{tbl:transductive}
\end{table}

\subsection{Transductive Semi-supervised Learning}
Generally, semi-supervised learning may refer to either inductive learning or transductive learning. In inductive learning, the goal is to generalize correct labels for new unseen data. In transductive learning, the goal is infer the labels restricted to the given unlabeled data. 
Our previous experiments conducted on unseen validation set can be considered as inductive learning. In Table \ref{tbl:transductive} we show that our SESS is also effective in transductive learning on both datasets. Our SESS consistently outperforms the fully-supervised VoteNet under different proportions of labeled samples. This demonstrates that our proposed SESS is a general framework that is not specific to inductive or transductive solution.

\begin{figure}[t]
	\centering
	\begin{subfigure}[b]{0.31\textwidth}
		\includegraphics[width=\linewidth]{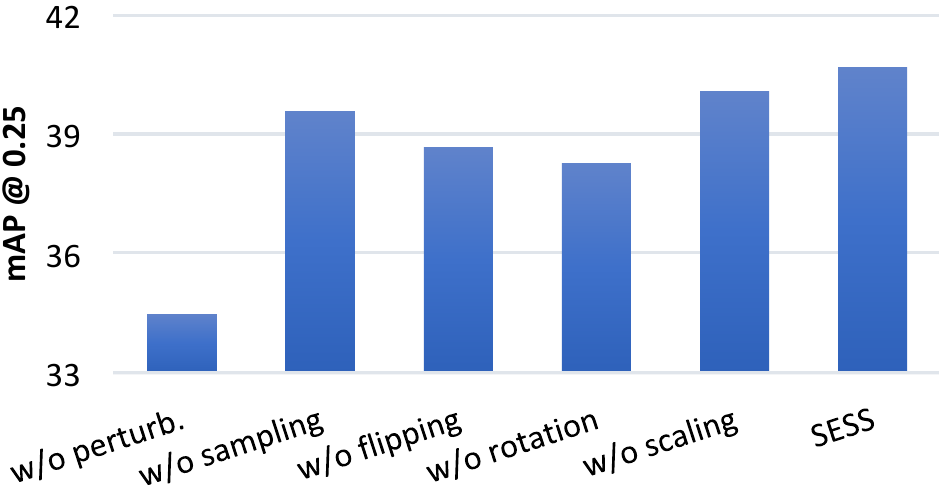}
		\caption{SUN RGB-D}
		\label{fig:ablation-perturbations-sunrgbd}
	\end{subfigure} 
	\begin{subfigure}[b]{0.31\textwidth}
		\includegraphics[width=\linewidth]{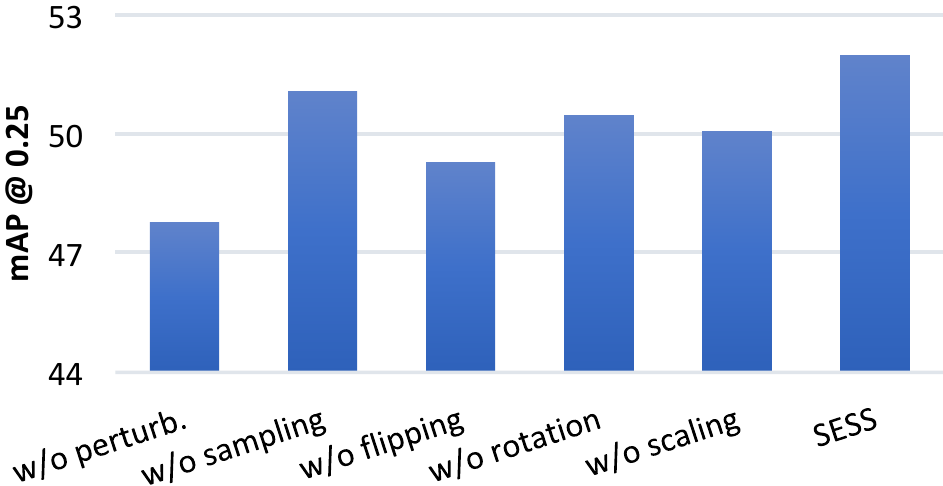}
		\caption{ScanNetV2}
		\label{fig:ablation-perturbations-scannet}
	\end{subfigure}
	\caption{Effects of different perturbations.}
	\label{fig:ablation-perturbations}
	\vspace{-0.1in}
\end{figure}

\subsection{Ablation Studies}
In this section, we explore the effects of perturbations and consistency losses. The training of ablation experiments is conducted on SUN RGB-D with 10\% labeled data and ScanNetV2 with of 30\% labeled data. The evaluation is on the corresponding validation set. 
\vspace{-0.1in}
\paragraph{Perturbations.} We study the effect of each perturbation by removing it from the framework, and report the performance after the removal. We also evaluate an extreme case that removes the perturbation scheme altogether. Figure \ref{fig:ablation-perturbations} illustrates the resultant performances. 
Obviously, the performance drops greatly on both datasets when the entire perturbation scheme is removed.
The effect may vary between the datasets for each individual perturbation. For example, the rotating perturbation contributes less to performance on ScanNet than SUN RGBD, as the bounding boxes of objects in ScanNet are axis-aligned. The scaling perturbation gives less improvement on SUN RGB-D than that on ScanNet. We suspect that this is because the partial scenes in SUN RGB-D are all with similar scales and thus are less sensitive to scaling perturbation. In contrast, the scales of the scenes in ScanNet are quite diverse. 
\vspace{-0.1in}
\paragraph{Consistency Losses.} We further investigate the effects of our three consistency losses by experimenting with different combinations. The comparison is reported in Table \ref{tab:ablation-consistencyloss}. From the perspective of individual consistency loss, the center-aware and class-aware consistency losses contribute more than the size-aware consistency loss. However, the combination of center-aware or class-aware with size-aware consistency loss helps to improve the performance to some extent. Finally, the integration of the three consistency losses gives us the best performance on both datasets.
It indicates that the requirement of representing the predicted bounding boxes with correct geometries (\ie center, size) as well as semantics (\ie class) regularizes the model towards a better performance.

\begin{table}[t]
	\centering
	\caption{Ablation study on consistency losses.}
	\vspace{-0.1in}
	\scalebox{0.85}{
		\begin{tabular}{c c c || c | c}
			\hline
			center & class & size & SUN RGB-D & ScanNetV2 \\ \hline 
			\cmark   & \xmark  & \xmark & 38.2 & 50.0    \\
			\xmark  &  \cmark  & \xmark & 39.2 & 50.2    \\
			\xmark & \xmark  &  \cmark  & 38.1 & 49.2  \\
			\xmark  &  \cmark &  \cmark & 40.3  & 50.7  \\
			\cmark& \xmark  &  \cmark & 38.9 &  50.5 \\
			\cmark& \cmark & \xmark  & 40.0 & 51.5    \\
			\cmark & \cmark & \cmark  & \textbf{40.7} &\textbf{52.0} \\\hline
	\end{tabular}}
	\label{tab:ablation-consistencyloss}
\end{table}

\subsection{Qualitative Results and Analysis}
\begin{figure*}[t]
	\centering
	\includegraphics[scale=0.51]{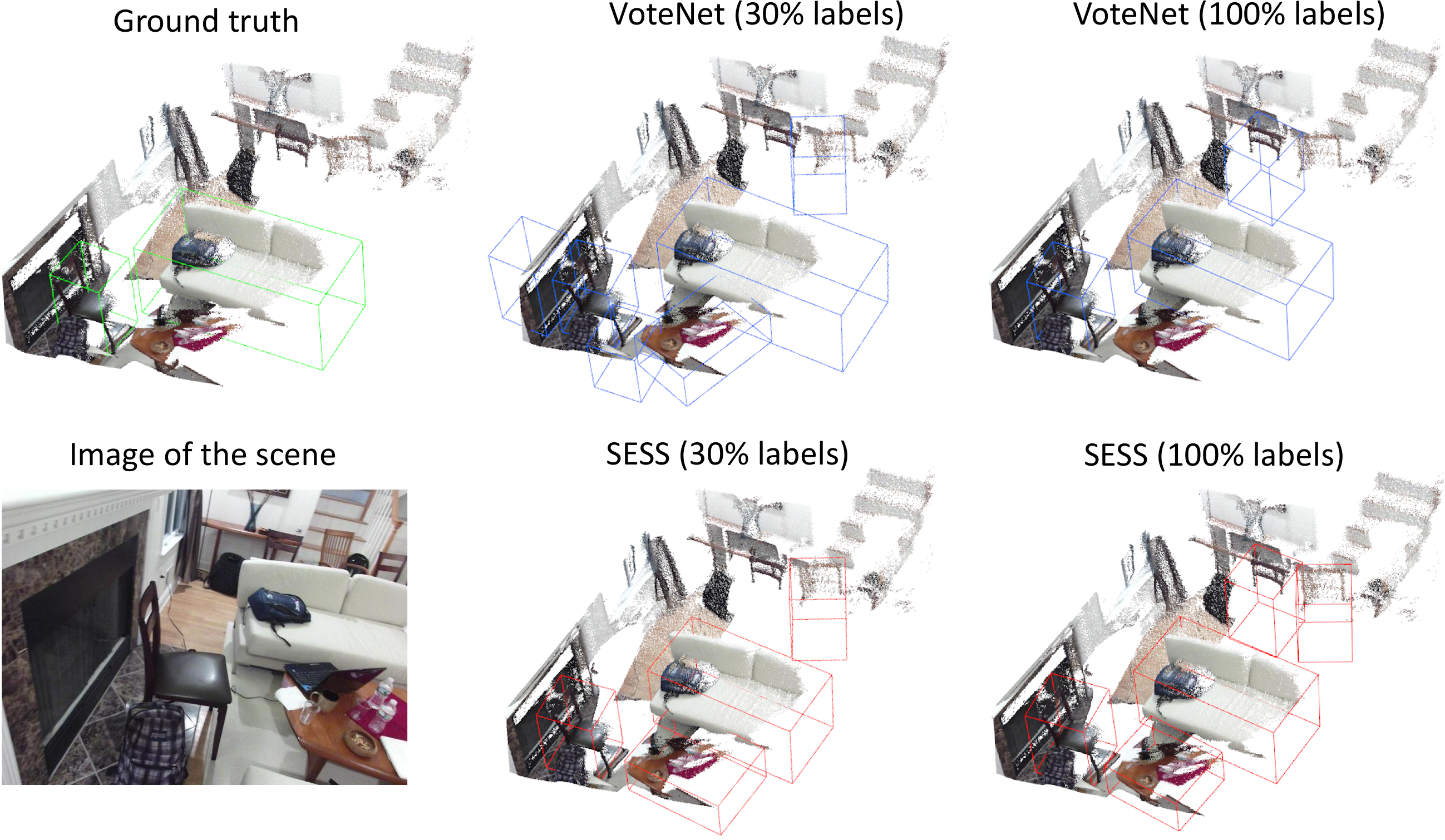}
	\caption{Qualitative comparison between the fully-supervised VoteNet and the proposed SESS on SUN RGB-D val set.}
	\label{fig:qualitative_sunrgbd}
\end{figure*}

\begin{figure*}[t]
	\centering
	\includegraphics[scale=0.55]{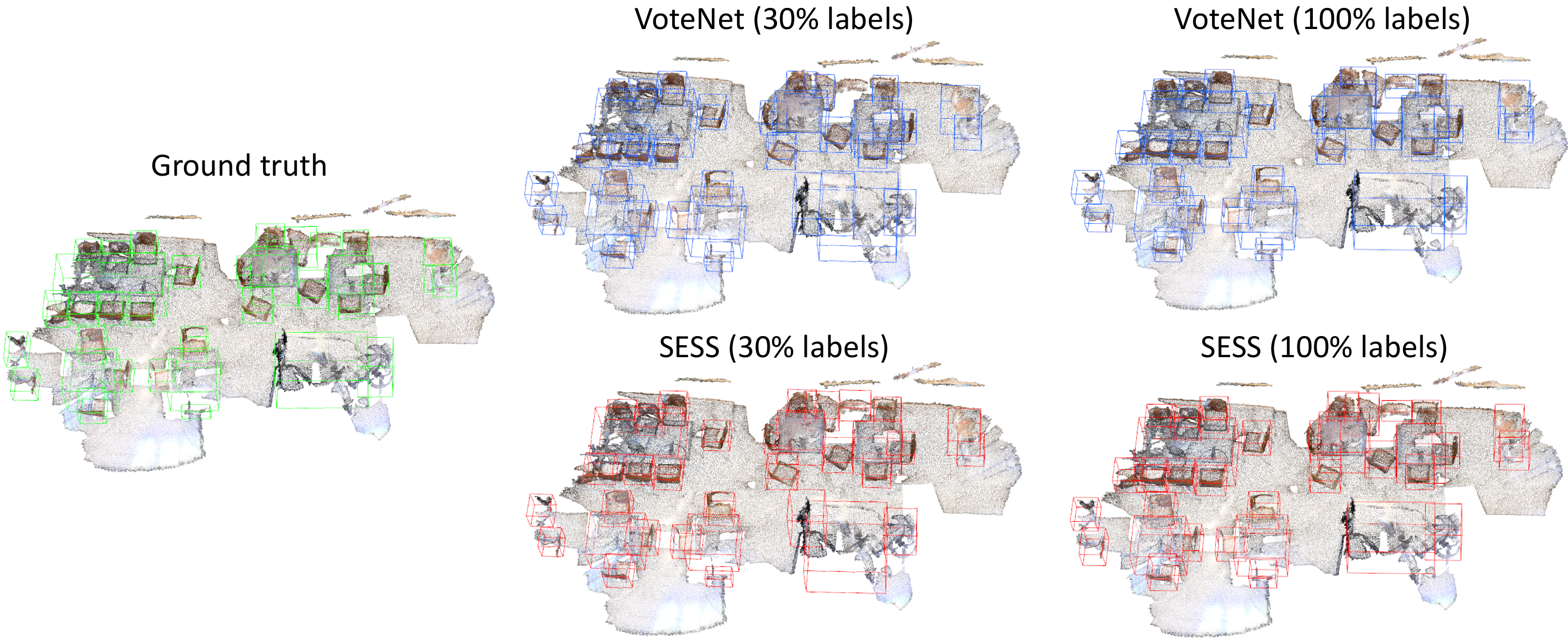}
	\caption{Qualitative comparison between the fully-supervised VoteNet and the proposed SESS on ScanNetV2 val set.}
	\label{fig:qualitative_scannet}
\end{figure*}

Figure \ref{fig:qualitative_sunrgbd} and Figure \ref{fig:qualitative_scannet} show the visualizations of the predictions by VoteNet and SESS with 30\% labeled training data and 100\% labeled training data on the ScanNet and SUN RGB-D scenes, respectively. 
As seen in Figure \ref{fig:qualitative_sunrgbd}, the partial scene obtained by single-view scanning in SUN RGB-D is very challenging, where some objects are partly visible but annotated with amodal ground-truth bounding boxes (\eg the ``sofa'' in Figure \ref{fig:qualitative_sunrgbd}). Surprisingly, both our method and the strongly supervised VoteNet successfully detect the target objects in such a challenging scene. 
Similar to the strongly supervised VoteNet, our SESS is able to detect more objects than those provided by the ground-truth annotations, such as the partial table in front of the sofa and the heavily occluded chairs behind the sofa. Our SESS gives more accurate predictions than VoteNet in terms of unannotated objects
with 30\% labeled data. We attribute this to the exploitation of unlabeled data in our proposal approach. Our SESS detects more unannotated objects when 100\% labeled data is used in training, and the predicted 3D bounding boxes are consistent with human perception.  

In contrast to the partial scenes in SUN RGB-D, the scenes in ScanNet are more complete and include larger areas with cluttered objects. An example is shown in Figure \ref{fig:qualitative_scannet}, this scene contains 7 tables and 27 chairs. Our SESS correctly recognizes the 7 tables and 26 chairs
with 30\% labeled data, while the strongly supervised VoteNet only detects 6 tables and 24 chairs correctly. 
We argue that the proposed consistency losses, which guide the model with encoded geometric and semantic information, contribute to the better localization of the 3D bounding boxes.
All 34 objects are completely detected with precise bounding boxes  
when our model is trained with 100\% labeled data.

\section{Conclusion}
In this paper, we propose SESS, a novel self-ensembling semi-supervised point cloud-based 3D object detection framework. It does not require a large amount of strong labels that are often difficult to obtain. 
Our SESS follows the Mean Teacher paradigm, where we design a perturbation scheme specific to point-based data and three consistency losses that are able to force the network to generate more accuracy detections.  
The experimental results on two real-world datasets validate the effectiveness and advantage of our SESS. And we experimentally show that our method is a general framework that can be applied in both inductive and transductive semi-supervised 3D object detection.

\paragraph{Acknowledgements.}
This research is supported in part by the National Research Foundation, Singapore under its International Research Centres in Singapore Funding Initiative. Any opinions, findings and conclusions or recommendations expressed in this material are those of the author(s) and do not reflect the views of National Research Foundation, Singapore. It is also partially supported by the Singapore MOE Tier 1 grant R-252-000-A65-114.
{\small
	\bibliographystyle{ieee_fullname}
	\bibliography{egbib}
}

\newpage

\appendix
In this appendix, we provide performance comparison between SESS and VoteNet with more diverse ratios of labeled data on the SUN RGB-D and ScaNetV2 val sets in Sec.~\ref{sec:additional_labelratio}. We also provide additional evaluation metric (\ie mAP@0.5 IoU) for both inductive and transductive semi-supervised learning in Sec.~\ref{sec:additional_evaluation}. In Sec.~\ref{sec:perclass_evaluation}, we report per-class average precision on the SUN RGB-D and ScanNetV2 val set. Finally, more qualitative results are shown in Section~\ref{sec:qualitative}. 

\section{Additional Label ratios}\label{sec:additional_labelratio}
\begin{figure}[h]
	\centering
	\begin{subfigure}[b]{0.45\textwidth}
		\includegraphics[width=\linewidth]{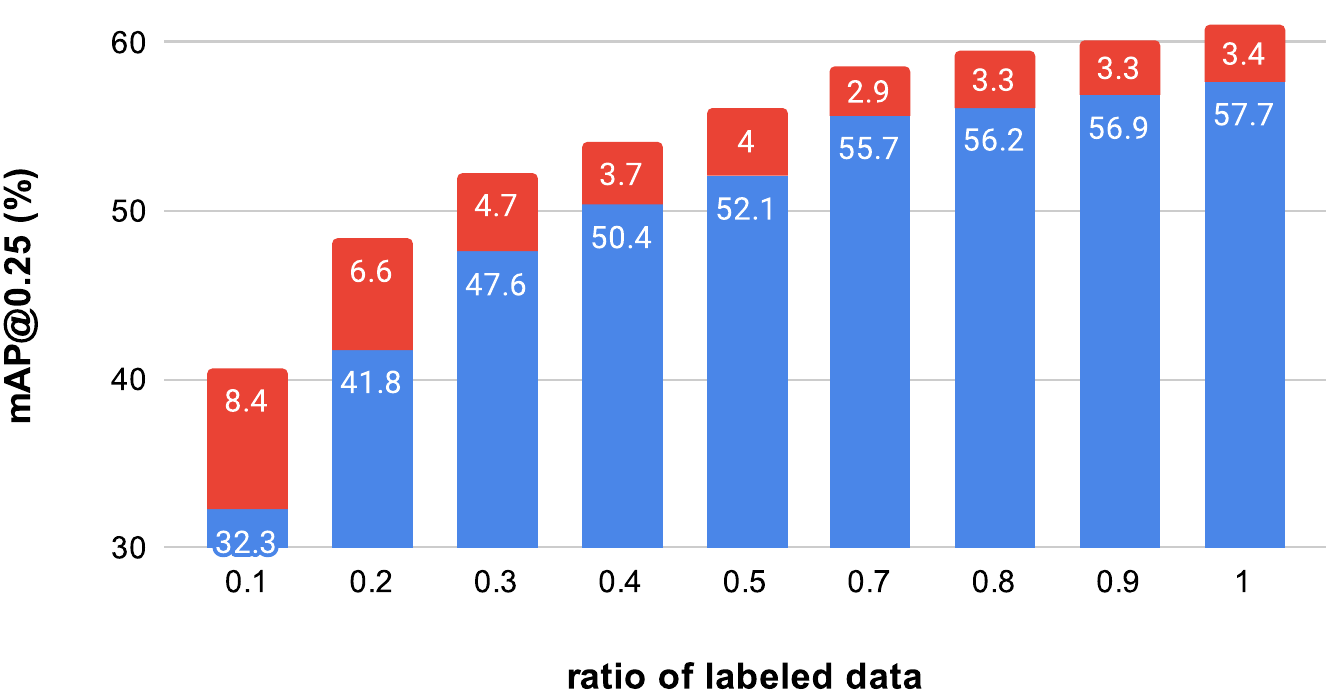}
		\caption{SUN RGB-D}
		\label{fig:comparison-sunrgbd}
	\end{subfigure} 
	\begin{subfigure}[b]{0.45\textwidth}
		\includegraphics[width=\linewidth]{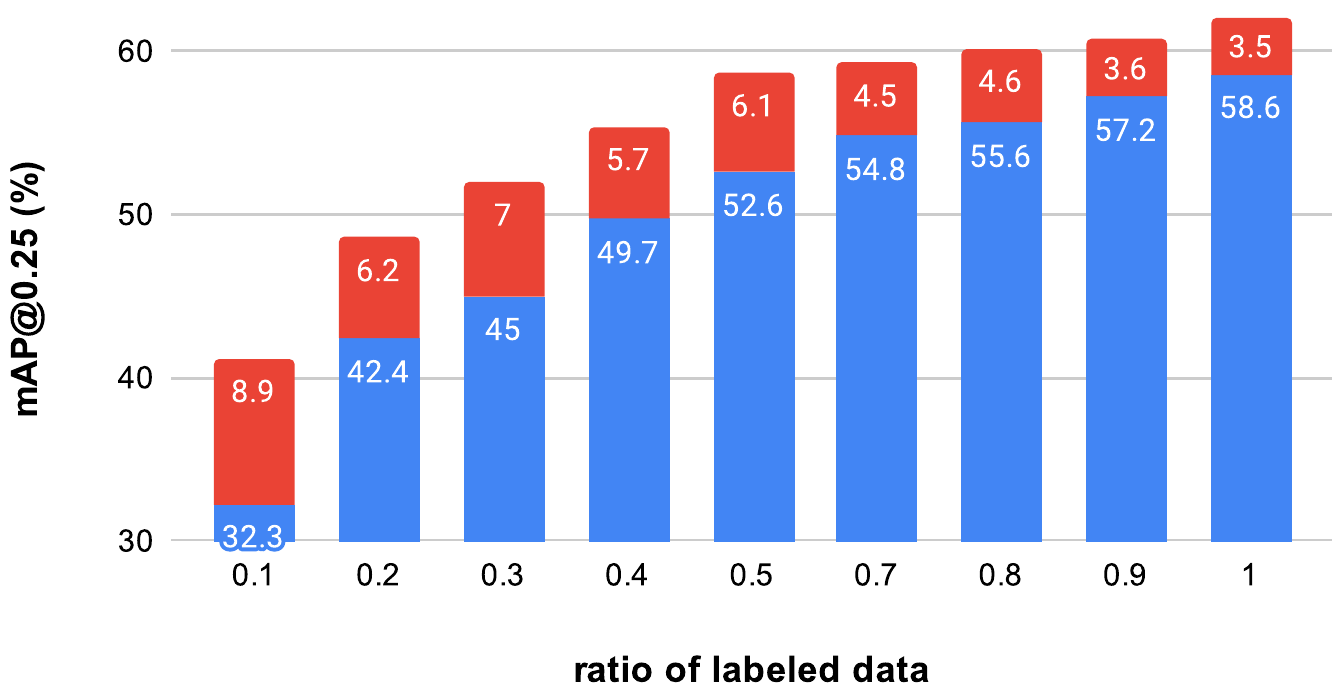}
		\caption{ScanNetV2}
		\label{fig:comparison-scannet}
	\end{subfigure}
	\caption{Comparison to VoteNet with more ratios of labeled data on the SUN RGB-D and ScanNetV2 val sets. The blue columns denote the performances of VoteNet, and the red columns denote the improved performance of SESS over VoteNet.}
	\label{fig:more_comparison}
\end{figure}

More ratios (\ie 80\% and 90\%) of labeled data are included in the performance comparison of our SESS to the fully-supervised VoteNet on two datasets. The comparison results are illustrated in Figure \ref{fig:more_comparison}.
As can be seen from the figures, the performance margin (compared to the performance of using 100\% labeled data) becomes smaller when the ratio of labeled data increases. This is because the same type of scenes (\eg classrooms) share similar layout and/or objects, and thus the contribution of new labeled data to model training might be minor when similar types of samples/scenes have been seen by the model.

\section{Additional Evaluation Metric}\label{sec:additional_evaluation}
We additionally evaluate mean average precision with an IoU threshold of 0.5 on the SUN RGB-D and ScanNetV2 for both inductive (see Table~\ref{tbl:inductive-map0.5}) and transductive (see Table~\ref{tbl:transductive-map0.5}) semi-supervised 3D object detection. 
Consistent with the evaluation at an IoU threshold of 0.25, our SESS significantly outperforms the fully supervised VoteNet under different ratios of labeled data for both inductive and transductive learning.

\begin{table}[t]
	\centering
	\caption{Inductive leaning on SUN RGB-D and ScanNetV2 val sets compared with the fully supervised VoteNet, evaluated by mAP@0.5 IoU. The percentage indicates the ratio of labeled data for training.}
	\scalebox{0.74}{
		\begin{tabular}{l ||l|l l l l l l l}
			\hline
			Dataset & Model  & 10\% & 20\% & 30\% & 40\% & 50\% & 70\%  & 100\% \\
			\hline
			\multirow{2}{*}{\small{SUNRGB-D}}
			& VoteNet & 10.6 & 14.7 & 23.3 & 25.6 & 27.2 & 30.0 & 31.1\\ \cline{2-9}
			&\textbf{SESS} & \textbf{14.4} & \textbf{20.6} &\textbf{28.5} & \textbf{29.0} & \textbf{30.6}  & \textbf{33.4} & \textbf{37.3}\\
			\hline \hline
			\multirow{2}{*}{ScanNetV2}
			& VoteNet & 11.9 & 21.2 & 22.5 & 27.7 & 28.9 & 30.9 & 33.5\\ \cline{2-9}
			&\textbf{SESS} & \textbf{18.6} & \textbf{26.9} &  \textbf{27.4} & \textbf{31.5} & \textbf{34.2} & \textbf{35.5} & \textbf{38.8}\\
			\hline
		\end{tabular}
	}
	\label{tbl:inductive-map0.5}
\end{table}

\begin{table}[t]
	\centering
	\caption{Transductive leaning on SUN RGB-D and ScanNetV2 unlabeled training sets compared with the fully supervised VoteNet, evaluated by mAP@0.5 IoU. The percentage indicates the ratio of labeled data for training.}
	\scalebox{0.83}{
		\begin{tabular}{l ||l|l l l l l l}
			\hline
			Dataset & Model  & 10\% & 20\% & 30\% & 40\% & 50\% & 70\%  \\
			\hline
			\multirow{2}{*}{\small{SUNRGB-D}}
			& VoteNet & 10.3 & 15.3 & 23.4  & 25.5 & 25.0 & 29.9\\ 
			&\textbf{SESS} & \textbf{15.8} & \textbf{20.1} & \textbf{27.4} & \textbf{27.2} & \textbf{29.2} & \textbf{36.7}\\
			\hline \hline
			\multirow{2}{*}{ScanNetV2}
			& VoteNet & 13.8 & 25.3 & 28.6 & 32.7 & 35.2 & 38.3 \\ 
			&\textbf{SESS} & \textbf{23.2} & \textbf{31.3} &  \textbf{34.3} & \textbf{37.6} & \textbf{41.6} & \textbf{42.6}\\
			\hline
		\end{tabular}
	}\label{tbl:transductive-map0.5}
\end{table}

\begin{table*}[t]
	\centering
	\caption{Per-class AP@0.25 IoU on SUN RGB-D val set, with 100\% training samples. The upper table lists the results obtained by five fully-supervised methods, and the lower table lists the results of our proposed semi-supervised method.}
	\scalebox{0.9}{
		\begin{tabular}{l|l l l l l l l l l l | l}
			\hline
			Method & bathtub & bed & bookshelf & chair & desk & dresser & nightstand &  sofa & table  & toilet  & mAP  \\
			\hline
			DSS  & 44.2 & 78.8 & 11.9 & 61.2 & 20.5 & 6.4 & 15.4 & 53.5 & 50.3 & 78.9 & 42.1 \\
			COG  & 58.3 & 63.7 & 31.8 & 62.2 & \textbf{45.2} & 15.5 & 27.4 & 51.0 & 51.3 & 70.1 & 47.6 \\
			2D-driven & 43.5 & 64.5 & 31.4 & 48.3 & 27.9 & 25.9 & 41.9 & 50.4 & 37.0 & 80.4 & 45.1 \\
			F-PointNet& 43.3 & 81.1 & 33.3 & 64.2 & 24.7 & \textbf{32.0} & 58.1 & 61.1 & 51.1 & 90.9 & 54.0 \\
			VoteNet  &74.4 & 83.0 & 28.8 & 75.3 & 22.0 & 29.8 & 62.2 & 64.0 & 47.3 & 90.1 & 57.7  \\
			\hline \hline
			\textbf{SESS} & \textbf{76.9} & \textbf{84.8} & \textbf{35.4} & \textbf{75.8} & 29.3 & 31.3 & \textbf{66.9} & \textbf{66.4} & \textbf{51.8} & \textbf{92.3} & \textbf{61.1}\\
			\hline
		\end{tabular}
	}
	\label{tbl:perclass-sunrgbd}
\end{table*}

\begin{table*}[t]
	\centering
	\caption{Per-class AP@0.25 IoU on ScanNetV2 val set, with 100\% training samples. The upper table lists the results from two fully-supervised methods, and the lower table lists the results of our proposed semi-supervised method.}
	\scalebox{0.74}{
		\begin{tabular}{l|l l l l l l l l l l l l l l l l l l | l}
			\hline
			Method & cabin. & bed & chair & sofa & table & door & wind. & bkshf & pic. & cntr & desk & curt. & refrig.  & showr. & toilet & sink & bath & ofurn. & mAP  \\
			\hline
			3DSIS  & 19.8 & 69.7 & 66.2 & 71.8 & 36.1 & 30.6 & 10.9 & 27.3 & 0.0 & 10.0 & 46.9 & 14.1 & 53.8 & 36.0 & 87.6 & 43.0 & 84.3 & 16.2 & 40.2 \\
			VoteNet & 36.3  & 87.9  & \textbf{88.7} &  89.6  & 58.8 &  47.3  & 38.1 &  44.6 &  7.8 &  \textbf{56.1}  & 71.7  & \textbf{47.2}  & 45.4 &  57.1  & 94.9  & \textbf{54.7}  & 92.1 &  37.2 & 58.6\\
			\hline \hline
			\textbf{SESS} & \textbf{41.1} & \textbf{88.1} & 85.9 & \textbf{91.7} & \textbf{64.5} & \textbf{52.1} & \textbf{40.4} & \textbf{51.4} & \textbf{11.8} & 51.9 & \textbf{74.9} & 45.9 & \textbf{59.6} & \textbf{73.3} & \textbf{98.3} & 53.9 & \textbf{93.0} & \textbf{39.5} & \textbf{62.1}\\
			\hline
		\end{tabular}
	}
	\label{tbl:perclass-scannet}
\end{table*}

\section{Per-class Evaluation}\label{sec:perclass_evaluation}
We respectively report per-class average precision on 10 classes of SUN RGB-D and 18 classes of ScanNetV2 in Table \ref{tbl:perclass-sunrgbd} and \ref{tbl:perclass-scannet}, using all the training samples. Our SESS is superior than the fully supervised VoteNet on each class of SUN RGB-D and 14 classes of ScanNetV2 with the assistance of the proposed pertubation scheme and consistency losses.

\section{More Qualitative Results and Discussions}\label{sec:qualitative}
Figure \ref{fig:qualitative_sunrgbd} and \ref{fig:qualitative_scannet} demonstrate additional qualitative results on the SUN RGB-D and ScanNetV2 val datasets, respectively. As can be seen from the four examples in Figure \ref{fig:qualitative_sunrgbd}, the heavy occlusion (\eg the chairs at the back rows in the classroom), partial visibility (\eg the leftmost cabinet in the bedroom), and extreme sparsity (\eg the rightmost chair in the study space) make the detection on SUN RGB-D very difficult. Some of them are even hard for human to recognize without the reference of the associated RGB images, such as the leftmost chair in the second row in the classroom and the rightmost chair in the study space. Both VoteNet and our SESS fail to detect these extremely challenging objects that come with no or few representative points. However, it is interesting to see that our SESS successfully detect most of the objects in these challenging scenarios, including those unannotated objects such as the chairs in the back of the classroom, and the table in front of the bed in the bedroom. 

In Figure \ref{fig:qualitative_scannet}, we also show four more examples covering various scenarios on ScanNetV2 dataset. Objects with strong geometric cues (\eg table, chair, bed, desk \etc) are easy to detect
since both strongly supervised VoteNet and our SESS rely on only the geometric data (\ie XYZ coordinates). In contrast, objects without explicit geometric features (\eg door, picture, window) are difficult to recognize. Despite the challenge, our SESS is able to detect most of the difficult objects, such as bookshelves in the library and doors in the lounge. We argue that the proposed consistency losses, which encode not only geometric but also semantic information, guide the model to achieve better localization of the 3D bounding boxes.

\begin{figure*}[t]
	\centering
	\includegraphics[scale=0.75]{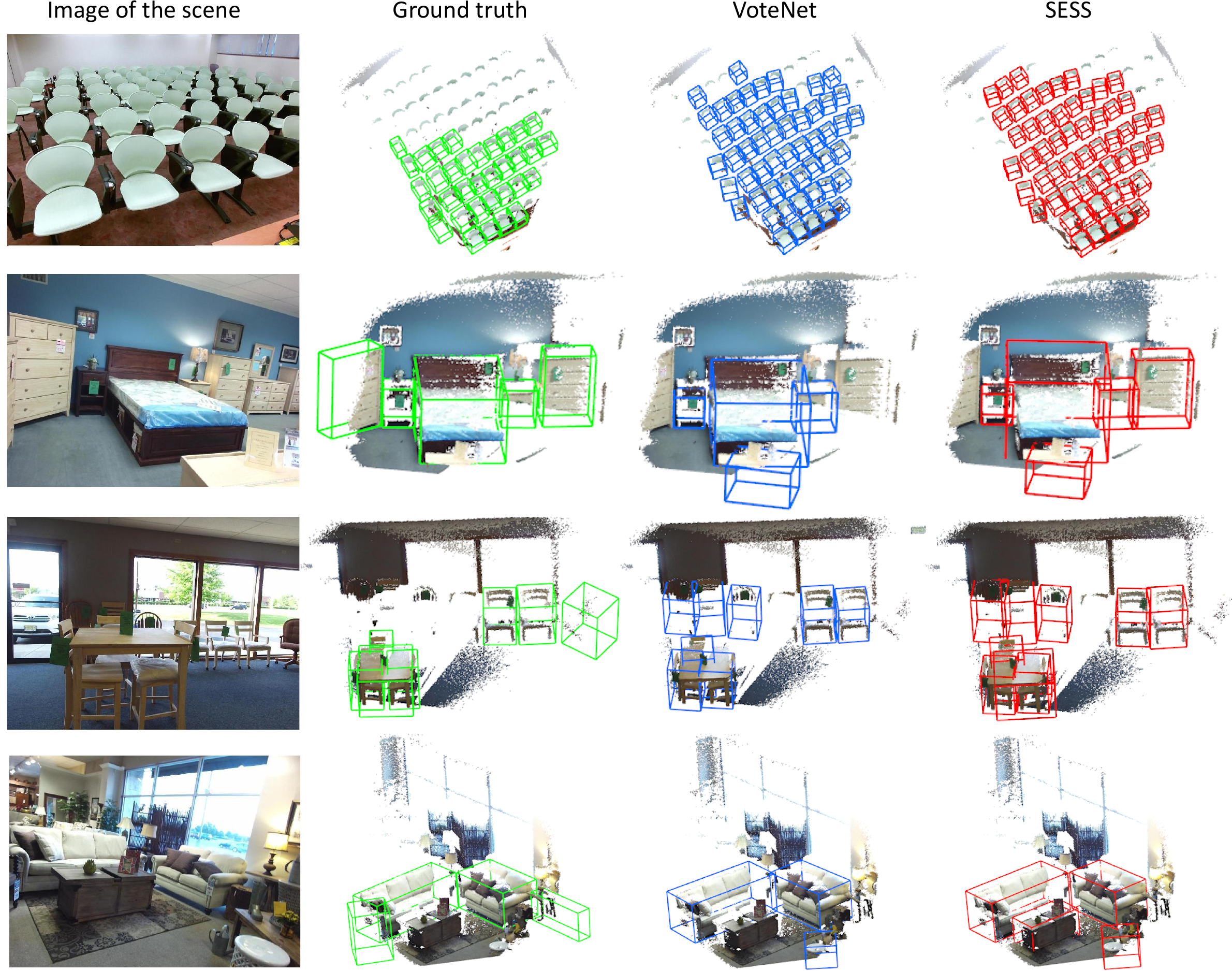}
	\caption{Additional Qualitative comparison between the fully-supervised VoteNet and the proposed SESS on SUN RGB-D \textit{val} set, using 100\% training samples. Four scene types are illustrated from the upper to bottom, they are \textit{classroom}, \textit{bedroom}, \textit{study space}, and \textit{living room}.}
	\label{fig:qualitative_sunrgbd}
\end{figure*}

\begin{figure*}[t]
	\centering
	\includegraphics[scale=1.2]{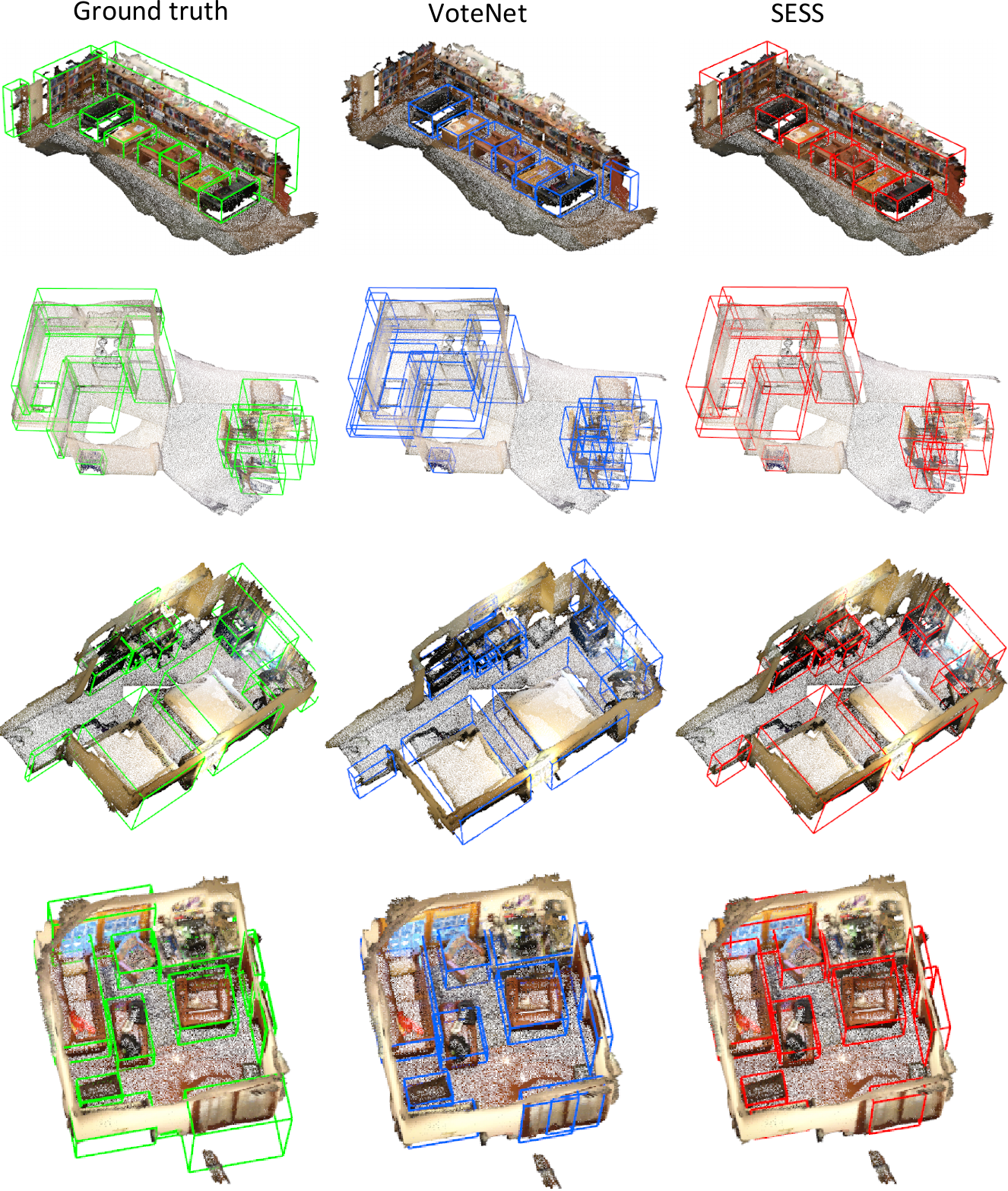}
	\caption{Additional Qualitative comparison between the fully-supervised VoteNet and the proposed SESS on ScanNetV2 \textit{val} set, using 100\% training samples. Four scene types are illustrated from the upper to bottom, they are \textit{library}, \textit{kitchen}, \textit{hotel}, and \textit{lounge}.}
	\label{fig:qualitative_scannet}
\end{figure*}

\end{document}